\documentclass[runningheads]{llncs}
\usepackage{graphicx}
\usepackage{amsmath,amssymb} % define this before the line numbering.
\usepackage{color}
\usepackage{bm}
\usepackage{float} 
\usepackage{subfigure}
\usepackage{caption}
\usepackage{multirow}
\usepackage{booktabs}

\usepackage[width=122mm,left=12mm,paperwidth=146mm,height=193mm,top=12mm,paperheight=217mm]{geometry}
\usepackage{cite}
\usepackage{hyperref}
\usepackage{xcolor}
\usepackage{url}

\let\llncssubparagraph\subparagraph
\let\subparagraph\paragraph
\usepackage[compact]{titlesec}
\let\subparagraph\llncssubparagraph
\titlespacing*{\section}{0pt}{2.5ex plus 1ex minus .0ex}{2.0ex plus .0ex}
\titlespacing*{\subsection}{0pt}{1.5ex plus 1ex minus .0ex}{1.5ex plus .0ex}
\titlespacing*{\subsubsection}{0pt}{0.5ex plus 1ex minus .0ex}{0.1ex plus .0ex}
\hypersetup{
  colorlinks=true,
  linkcolor=green!75!black,
  citecolor=green!75!black,
  urlcolor=blue,  % 根据需要调整网址颜色
}

\begin{document}
% \linenumbers
\pagestyle{headings}

\mainmatter
\def\ECCV24SubNumber{***}  % Insert your submission number here

\title{PKU-I2IQA: An Image-to-Image Quality Assessment Database for AI Generated Images} % Replace with your title

\titlerunning{PKU-I2IQA: An Image-to-Image Quality Assessment Database for AI Generated Images}
\authorrunning{J. Yuan, X. Cao, C. Li, F. Yang, J. Lin, X. Cao}

\author{Jiquan Yuan, Xinyan Cao, Changjin Li, Fanyi Yang, \\Jinlong Lin, \and
Xixin Cao\thanks{Corresponding author. Email: cxx@ss.pku.edu.cn}}
\institute{School of Software \& Microelectronics, Peking University, Beijing, China}

\maketitle

\begin{abstract}
As image generation technology advances, AI-based image generation has been applied in various fields and Artificial Intelligence Generated Content (AIGC) has garnered widespread attention. However, the development of AI-based image generative models also brings new problems and challenges. A significant challenge is that AI-generated images (AIGI) may exhibit unique distortions compared to natural images, and not all generated images meet the requirements of the real world. Therefore, it is of great significance to evaluate AIGIs more comprehensively. Although previous work has established several human perception-based AIGC image quality assessment (AIGCIQA) databases for text-generated images, the AI image generation technology includes scenarios like text-to-image and image-to-image, and assessing only the images generated by text-to-image models is insufficient. To address this issue, we establish a human perception-based image-to-image AIGCIQA database, named PKU-I2IQA. We conduct a well-organized subjective experiment to collect quality labels for AIGIs and then conduct a comprehensive analysis of the PKU-I2IQA database. Furthermore, we have proposed two benchmark models: NR-AIGCIQA based on the no-reference image quality assessment method and FR-AIGCIQA based on the full-reference image quality assessment method. Finally, leveraging this database, we conduct benchmark experiments and compare the performance of the proposed benchmark models. The PKU-I2IQA database and benchmarks will be released to facilitate future research on \url{https://github.com/jiquan123/I2IQA}.
\end{abstract}

\section{Introduction}
In recent years, Artificial Intelligence Generated Content (AIGC) has garnered widespread attention beyond computer science, and society has become interested in various content-generation products developed by major technology companies. Image generation technology\cite{goodfellow2014generative,kingma2013vae,ho2020ddpm}, in particular, has experienced rapid development and has had a profound impact. With the development of image generation technology, AI-based image generation techniques have been applied across various fields. Many excellent image-generative models have emerged, such as Midjourney\cite{r1}, Stable Diffusion\cite{r2}, Glide\cite{nichol2021glide},
Lafite\cite{zhou2022lafite}, DALLE\cite{dalle}, Unidiffuser\cite{bao2023unidiff}, Controlnet\cite{zhang2023addingcontrolnet}, \textit{etc.} 

However, the advancement of AI image-generative models has also brought about new problems and challenges. A significant challenge is that AI-generated images (AIGI) may exhibit unique distortions compared to natural images. Not all generated images meet the requirements of the real world, often necessitating processing, adjustment, refinement, or filtering before practical application. In contrast to common image content\cite{r3,c3,c4,c5,c6} (such as natural scene images, screen content images, graphic images, \textit{etc.}), which typically encounter common distortions like noise, blur, compression, \textit{etc.}, AIGIs may suffer from distinctive degradation such as unrealistic structures, irregular textures and shapes, and AI artifacts\cite{zhang2023perceptual,wang2023aigciqa2023}, \textit{etc.} Additionally, AIGIs may not correspond to the semantics indicated by text prompts\cite{r4,r5,zhang2023perceptual,wang2023aigciqa2023}. As AIGIs continue to be produced, evaluating the quality of these images has become a significant challenge. Previously, AIGC image quality assessment (AIGCIQA) relies on automatic measures like Inception Score (IS)\cite{r6}, Fréchet Inception Distance (FID)\cite{r7}, and CLIP Score\cite{hessel2021clipscore}, \textit{etc.} However, research\cite{r8} points out that current evaluation metrics may fall short of expressing human perception. Particularly in terms of FID and Clip Score, they may no longer effectively evaluate the state-of-the-art generative models.
\begin{table}[t]
\caption{An overview of the AIGCIQA database}
\centering
\begin{tabular}{lcccc}
\toprule
Database & AIGI model & Text prompt & Image prompt & AIGI \\ \hline
AGIQA-1K\cite{zhang2023perceptual}       & 2                   & 1080                 & -                      & 1080          \\ \hline
AGIQA-3K\cite{AGIQA-3K}       & 6                   & 300                 & -                      & 2982          \\ \hline
AIGCIQA2023\cite{wang2023aigciqa2023}   & 6                   & 100                  & -                     & 2400          \\ \hline
\textbf{PKU-I2IQA}             & 2                   & 200                  & 200                   & 1600          \\ 
\bottomrule
\end{tabular}
\label{your-table-label}
\end{table}

Unfortunately, research in the field of AIGCIQA remains in its nascent stages. Notable strides have been made, as evidenced by the establishment of dedicated AIGCIQA databases, such as AGIQA-1K\cite{zhang2023perceptual}, AGIQA-3K\cite{AGIQA-3K}, and AIGCIQA2023\cite{wang2023aigciqa2023}. These databases represent significant progress in the realm of AIGCIQA. However, they predominantly focus on images produced via text-to-image models, thereby overlooking the diversity inherent in AI image generation technologies, which include both text-to-image and image-to-image generative methods.
This oversight highlights a critical gap in the current research landscape, underscoring the need for dedicated databases catering to image-to-image scenarios, as well as more comprehensive databases that encompass a broader range of AI-generated image scenarios. The establishment of such databases is imperative to enable a more holistic assessment for AIGC image quality.
Another issue pertains to the human perception-based approach utilized in the existing text-to-image AIGCIQA databases. The absence of reference images in these databases potentially introduces a bias in the human perception scores obtained from subjective experiments. Conversely, the establishment of image-to-image AIGCIQA databases, which utilize prompt images as references, could significantly mitigate this bias. This approach promises a more accurate and reliable collection of human annotations, paving the way for more balanced and objective evaluations in the field of AIGCIQA.

\begin{figure}[t]
\centering %表示居中
\includegraphics[height=8.664cm,width=10cm]{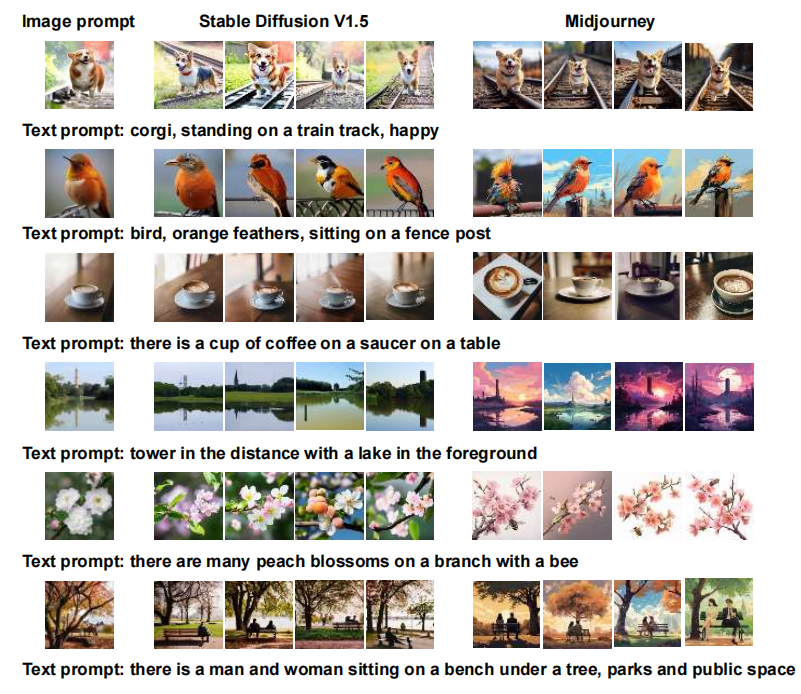}
% [height=4.5cm]表示高度
%[width=9.5cm]表示宽度
%{111.eps}表示eps格式的图片，名为111
\caption{Various scenes and styles of images sampled from the PKU-I2IQA database generated by Midjourney\cite{r1} and Stable Diffusion V1.5\cite{r2}.}
%图片的名称
\label{fig1}
%图片的标签，用于文章中的引用，注意到标签的数字与实际文章显示的数字可能不同
\end{figure}

To address the above issues, we first establish a human perception-based image-to-image database for AIGCIQA, named PKU-I2IQA. To the best of our knowledge, this is the first human perception-based image-to-image AIGCIQA database. Specifically, we select 200 categories from the well-known large-scale image database ImageNet\cite{russakovsky2015imagenet} in the field of computer vision. Subsequently, we collect corresponding images from the high-resolution image website Pixabay\cite{Pixabay} based on the selected categories to serve as image prompts for image-to-image generative models. These prompts include images of various scenes, such as animals, plants, furniture, and natural landscapes, \textit{etc.} We employ two popular image-to-image generative models Midjourney\cite{r1} and Stable Diffusion V1.5\cite{r2} as the AIGI models to generate images. For each image prompt, we generate four images randomly for each generative model. Therefore, the constructed PKU-I2IQA database comprises a total of 1600 images (\text{4 images} $\times$ \text{2 models} $\times$ \text{200 image prompts}) corresponding to 200 image prompts. We conduct a well-organized subjective experiment to collect quality labels for AIGIs and then conduct a comprehensive analysis of the PKU-I2IQA database. Table 1 compares the PKU-I2IQA database with existing AIGCIQA databases.

Different from previous works\cite{zhang2023perceptual, AGIQA-3K,wang2023aigciqa2023}, as the database is constructed using images generated by text-to-image models, there is no involvement of reference images when training and testing with deep learning models, which is corresponding to the no-reference image quality assessment method (NR-IQA) in image quality assessment. In contrast, the images in the PKU-I2IQA database are generated by image-to-image generative models using both image prompts and text prompts. Therefore, during training and testing, we can utilize image prompts as reference images which allows for a more accurate evaluation.
Depending on whether image prompts are provided as reference images during training and testing, we propose two benchmark models for AIGC image quality assessment: NR-AIGCIQA based on the no-reference image quality assessment (NR-IQA) method and FR-AIGCIQA based on the full-reference image quality assessment (FR-IQA) method. Finally, leveraging this database, we conduct benchmark experiments and compare the performance of the proposed benchmark models. The main contributions of this paper can be summarized as follows:

$\bullet$  We establish the first human perception-based image-to-image database for AIGCIQA, named PKU-I2IQA.

$\bullet$  We propose two benchmark models for AIGCIQA: NR-AIGCIQA based on the NR-IQA method and FR-AIGCIQA based on the FR-IQA method. 

$\bullet$  We conduct benchmark experiments and compare the performance of the proposed benchmark models on the PKU-I2IQA database.

\section{Related Work}
\subsubsection{Image Quality Assessment.}In the past few years, researchers have proposed numerous Image Quality Assessment (IQA) methods. IQA methods can be categorized into FR-IQA methods\cite{8063957,lao2022attentions,fr3,chen2021full} and NR-IQA methods \cite{nr-iqa,liu2017rankiqa,MetaIQA,yang2022maniqa,cnriqa,bmsb,mdaqa}, depending on whether a reference image is used during the prediction process. Full-reference methods often achieve higher prediction accuracy compared to no-reference methods, as the inclusion of a reference image allows the computer to extract more effective features during the prediction process. Many classical image quality assessment models initially employ methods based on manually extracted features\cite{nr-iqa,r19,r20}. However, with the rapid development of convolutional neural networks, methods based on deep learning for feature extraction\cite{8063957,lao2022attentions,fr3,chen2021full,liu2017rankiqa,MetaIQA,yang2022maniqa,cnriqa} have led to significant performance improvements. As a branch of image quality assessment, AIGC image quality assessment still requires further research. Previously, AIGCIQA relies on automatic measures like Inception Score (IS)\cite{r6}, Fréchet Inception Distance (FID)\cite{r7}, and CLIP Score\cite{hessel2021clipscore}, \textit{etc.} Recently, Mayu Otan \textit{et al.}\cite{r8} from the Japanese internet giant Cyber Agent conduct a detailed investigation and experiments on evaluation metrics for AIGCIQA. They find that current evaluation metrics are limited to express human perception, especially in terms of FID\cite{r7} and Clip Score\cite{hessel2021clipscore}, and are unable to evaluate the state-of-the-art generative models. Zhang \textit{et al.}\cite{zhang2023perceptual} establish the first human perception-based image-to-image database for AIGCIQA, named AGIQA-1K. It consists of 1,080 AIGIs generated by 2 diffusion models\cite{r2}. Through well-organized subjective experiments, human subjective perception evaluations of AIGIs are introduced to collect quality labels for AIGIs. Benchmark experiments are then conducted to evaluate the performance of the current IQA models\cite{he2016resnet,bmsb,mdaqa}. Li \textit{et al.} \cite{AGIQA-3K} consider six representative generative models and build the most comprehensive AIGI subjective quality database AGIQA-3K. This is the first database that covers AIGIs from GAN/auto regression/diffusion-based model altogether. Wang \textit{et al.}\cite{wang2023aigciqa2023} establish a large-scale AIGCIQA database, named AIGCIQA2023. They utilize 100 prompts and generate over 2000 images based on six state-of-the-art text-to-image generative models\cite{bao2023unidiff,dalle,zhou2022lafite,r2,nichol2021glide,zhang2023addingcontrolnet}. A well-organized subjective experiment is conducted on these images to evaluate human preferences for each image from the perspectives of quality, authenticity, and text-image correspondence. Finally, they perform benchmark experiments on this large-scale database to evaluate the performance of several state-of-the-art IQA models\cite{cnriqa,simonyan2014vgg,he2016resnet,8063957}. While these efforts have advanced the development of AIGCIQA, there are still issues to address, such as how to cover AIGC image generation in various scenarios as comprehensively as possible and how to introduce reference images into the AIGCIQA methods to enhance model performance. 
\subsubsection{Visual Backbone.}Visual Backbone Networks are fundamental and crucial components in computer vision, employed for feature extraction and representation in image processing tasks. These network models typically consist of multiple layers and modules designed to extract and represent features from input images, supporting various computer vision tasks such as object detection, image classification, semantic segmentation, \textit{etc.} In the last decade, deep learning has seen remarkable progress, especially after the introduction of ImageNet\cite{russakovsky2015imagenet} by Fei-Fei Li and her colleagues at Stanford University. This has significantly advanced deep learning's role in various computer vision tasks. We've seen the development of multiple visual backbone models, such as CNN-based ones like VGG\cite{simonyan2014vgg}, GoogleNet\cite{szegedy2015google}, ResNet\cite{he2016resnet}, and transformer-based ones like ViT\cite{dosovitskiy2020vit}, Swin Transformer\cite{liu2021swin}, \textit{etc.} In this paper, we employ several backbone network models pre-trained on the ImageNet\cite{russakovsky2015imagenet} as feature extraction networks. These networks are utilized to extract features from input images, and we evaluate the performance of different backbone network models.

\section{Database Construction and Analysis}
\subsection{AIGI Collection}

To ensure the diversity of the generated content, we select 200 categories from the famous large-scale image database ImageNet\cite{russakovsky2015imagenet} in the field of computer vision. Subsequently, we collect corresponding images from the high-resolution image website Pixabay\cite{Pixabay} based on the selected categories to serve as image prompts for image-to-image generative models. \textbf{It is explicitly stated that we use the royalty-free images from this website}. These prompts include images of various scenes such as animals, plants, furniture, and natural landscapes, \textit{etc.} Due to the varied resolutions of the collected prompt images from Pixabay, we standardize their resolution to $512\times512$, while preserving information about the image categories and scenes. This standardization involved resizing and cropping the images.

We employ two popular image generative models Midjourney\cite{r1} and Stable Diffusion V1.5\cite{r2} as our AIGI generative models. We first use Clip\cite{radford2021clip} to perform reverse deduction to obtain text prompts from image prompts. Subsequently, based on the image prompts, text prompts, and the specified parameters, we obtain the generated images with a resolution of $512\times512$.
For each image prompt, we generate four images randomly for each generative model. Consequently, our constructed PKU-I2IQA database comprises a total of 1600 images (\text{4 images} $\times$ \text{2 models} $\times$ \text{200 image prompts}), corresponding to 200 image prompts. Various scenes and styles of images sampled from the PKU-I2IQA database are shown in Fig.1.

\begin{figure}[t]
\centering %表示居中
\includegraphics[height=6cm,width=11.33cm]{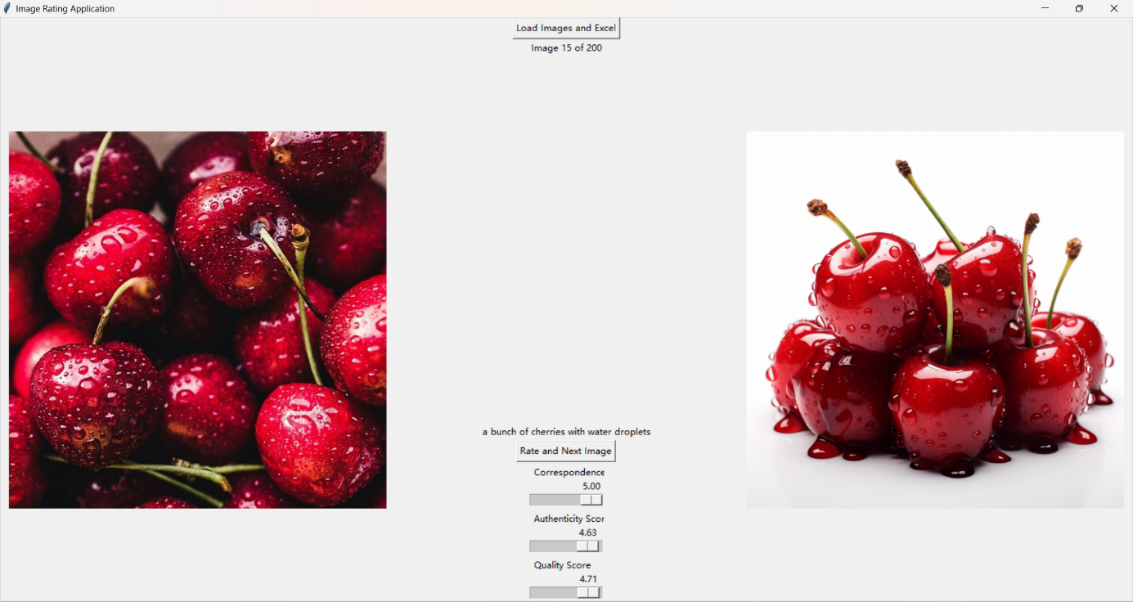}
\caption{An example of the subjective evaluation interface. Evaluators can assess the quality of AIGIs by comparing the reference image on the left with the to-be-evaluated AIGIs on the right. They can use the sliders below to record the text-image correspondence score, authenticity score, and quality score.}
%图片的名称
\label{fig2}
%图片的标签，用于文章中的引用，注意到标签的数字与实际文章显示的数字可能不同
\end{figure}

\subsection{Subjective Experiment}
To evaluate the image quality of the PKU-I2IQA database and obtain Mean Opinion Scores (MOSs), subjective experiments are conducted following the guidance of ITU-R BT.500-14\cite{ITUR}. Following previous work\cite{wang2023aigciqa2023}, evaluators are asked to express their preferences for the displayed AIGIs from three aspects: quality, authenticity, and text-image correspondence. Quality score is assessed based on clarity, color, brightness, and contrast of AI-generated images, along with sharpness of contours, detail richness, and overall aesthetic appeal.  Authenticity score focuses on whether the AI-generated images looks real and  whether evaluators could distinguish that the images are generated by AIGI generative models or not. Text-image correspondence scores refers to the matching degree between the generated images and the text prompts.

We employ a Python Tkinter-based graphical interface to display AIGIs in their native $512\times512$ resolution on the computer screen in a random sequence, as illustrated in Fig.2. Using this interface, evaluators rate AIGIs on a 0 to 5 scale with 0.01 increments. Unlike prior studies\cite{zhang2023perceptual,wang2023aigciqa2023}, we integrates image prompts as reference images into the graphical interface. This enables evaluators to conduct more accurate evaluation by directly comparing these images with the AIGIs under review. 

Twenty graduate students participate in our experiment, which is divided into eight stages to keep each evaluation session around an hour. In each stage, evaluators need to evaluate 200 AIGIs.

	\begin{figure}[t]%%图,[htbp]是浮动格式
	\subfigure[]{
	\includegraphics[width=6cm,height=2.58cm]{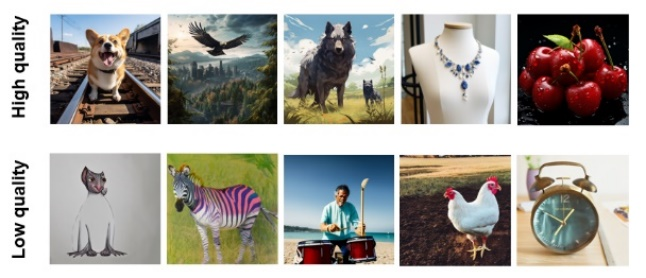} \label{Fig.3(a)}
}
	\hspace{2mm}
	\subfigure[]{
	\includegraphics[width=6cm,height=3.10cm]{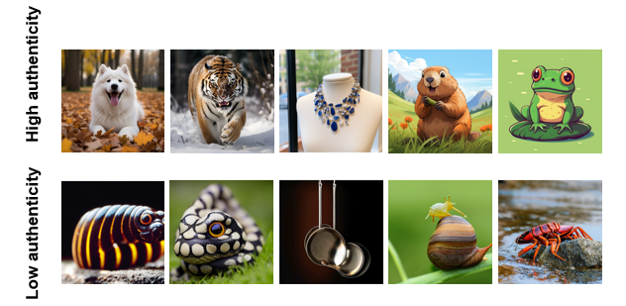} \label{Fig.3(b)}
}

%% 有回车	
	\hspace{2mm}
        \centering  
	\subfigure[]{
		\includegraphics[width=5.35cm,height=3.42cm]{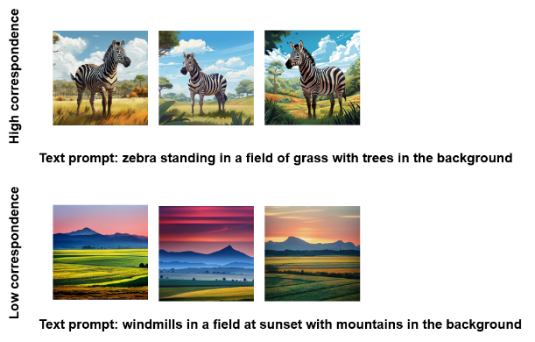} \label{Fig.3(c)}
	}
	
	\caption{Illustration of the images from the perspectives of quality, authenticity, and text-image correspondence. (a) 5 high-quality AIGIs and 5 low-quality AIGIs. (b) 5 high-authenticity AIGIs and 5 low-authenticity AIGIs. (c) 3 high text-image correspondence AIGIs and 3 low text-image correspondence AIGIs.}
	\end {figure}

\subsection{Data Processing}
After the subjective experiments, we collect ratings from all evaluators who participate in this experiment. Following the guidelines of ITU-R BT.500-14\cite{ITUR}, we calculate the mean and standard deviation of the subjective ratings for the same image within the same test group using the following formula:
\begin{align}
    \mu_j=\frac{1}{N}\sum_{i=1}^N r_{ij} \label{Eq.1}  \\
    S_j=\sqrt{\sum_{i=1}^N \frac{(\mu_j-r_{ij})^2}{N-1}} \label{Eq.2} %% 
\end{align}

\begin{figure}[t]
\centering
\subfigure[]{
\includegraphics[width=4.00cm,height=3.57cm]{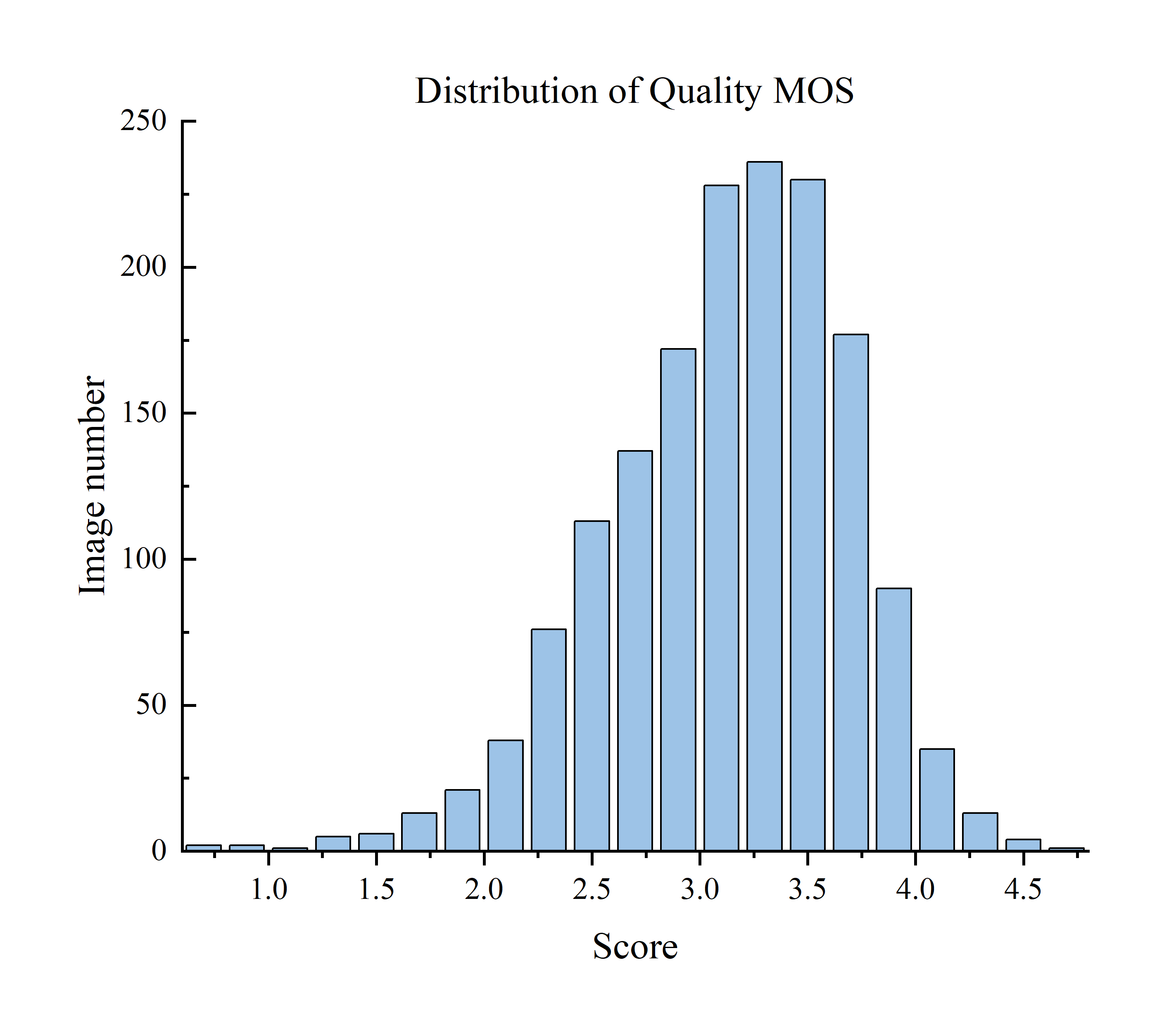} \label{1}
}
\quad
\subfigure[]{
\includegraphics[width=4.00cm,height=3.52cm]{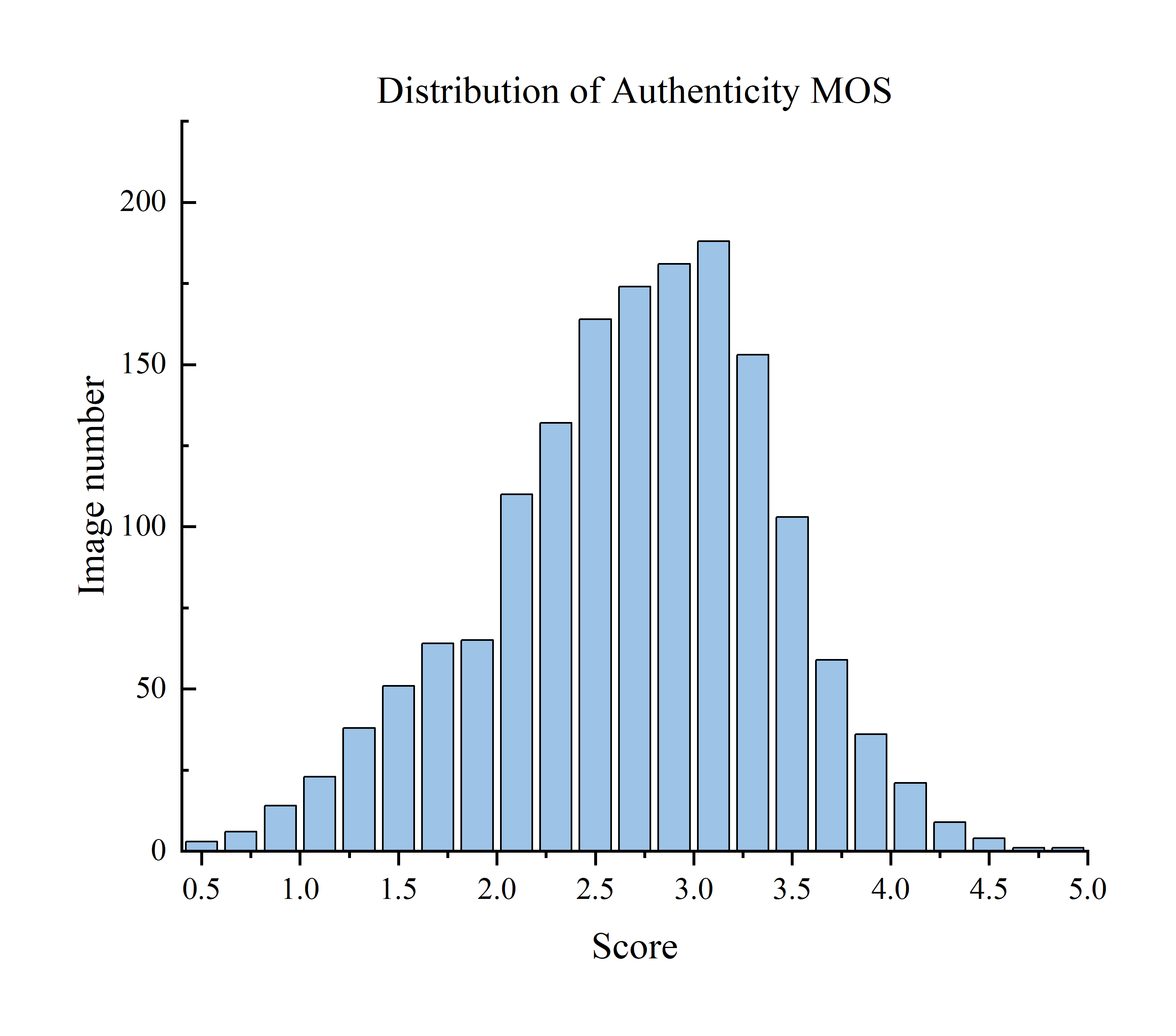} \label{2} 
}
\quad
\subfigure[]{
\includegraphics[width=4.00cm,height=3.52cm]{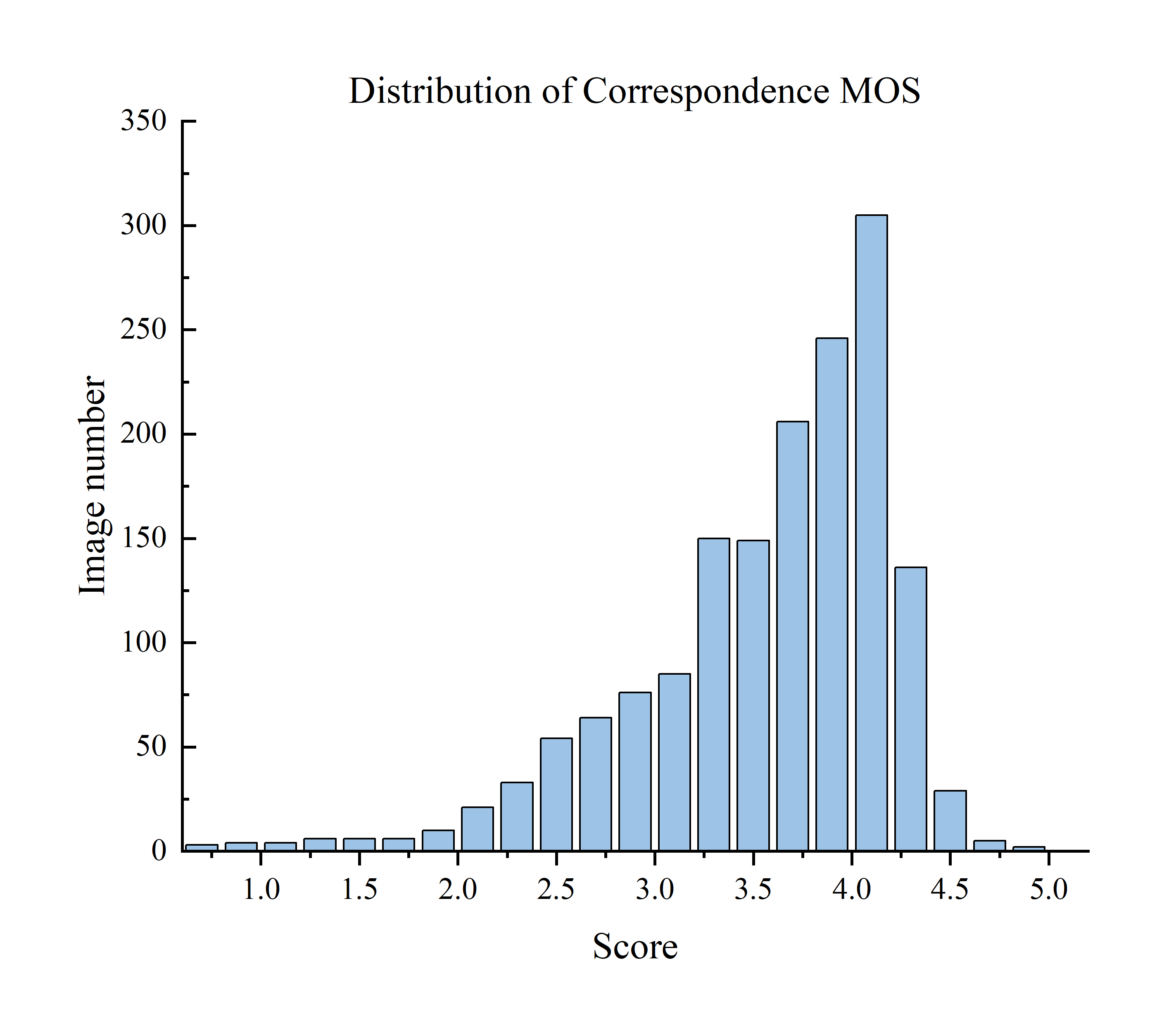}\label{3}
}
\quad
\subfigure[]{
\includegraphics[width=4.00cm,height=3.52cm]{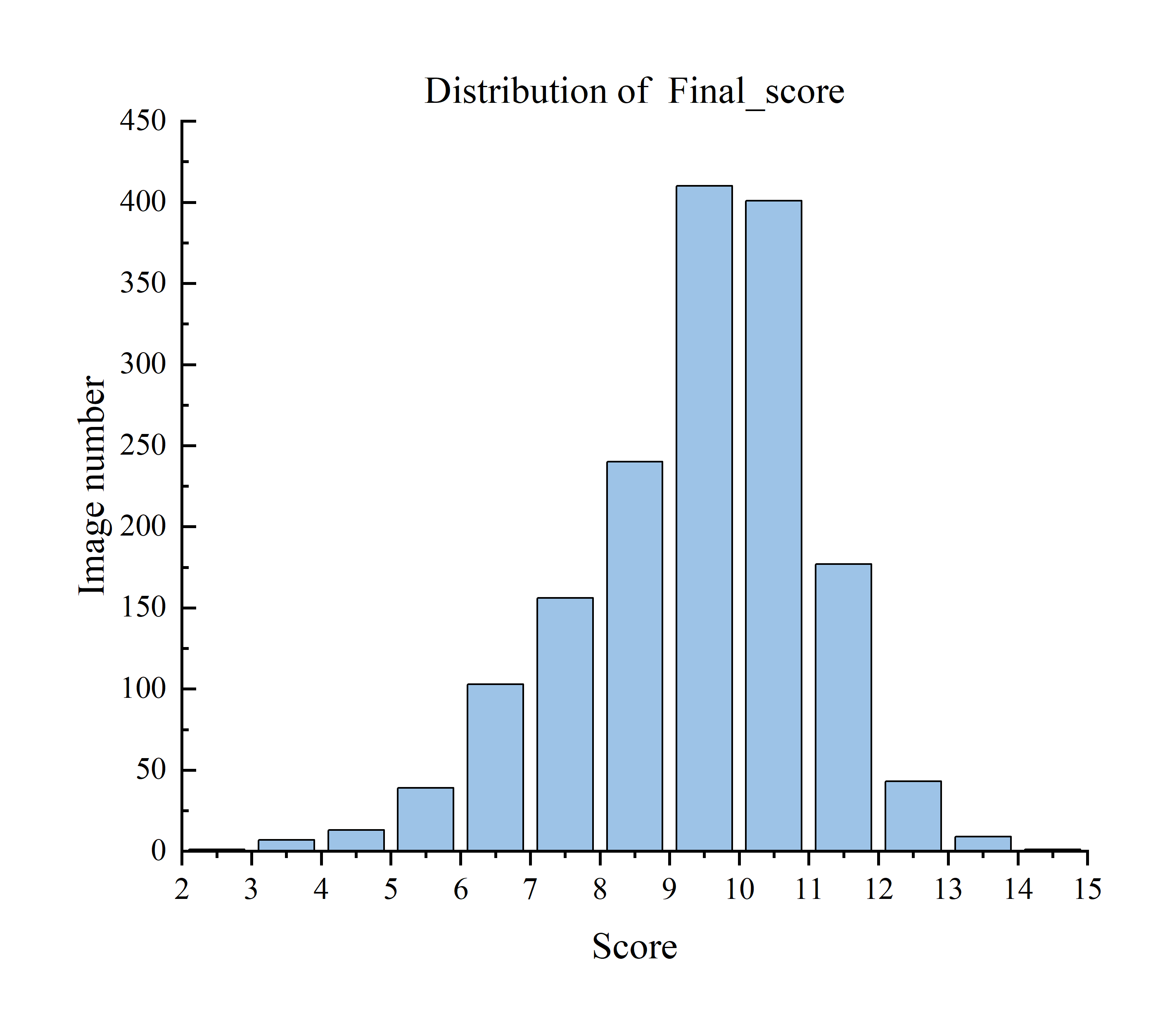}\label{4}
}
\caption{(a) MOSs distribution of quality score. (b) MOSs distribution of authenticity
score. (c) MOSs distribution of correspondence score. (d) Distribution of the Final\_score.}
\end{figure}

The notation $r_{ij}$ represents the score of the $i_{th}$ observer for the $j_{th}$ generated image, where $N$ denotes the total number of evaluators. When presenting the test results, all average scores should be accompanied by a relevant confidence interval, which derives from the standard deviation and the sample size. As recommended by ITU-R BT.500-14\cite{ITUR}, we employ a $95\%$ confidence interval $( \mu_j + \epsilon_j, \mu_j - \epsilon_j )$, where $\epsilon_j$ is computed using the following formula:
\begin{align}
    \epsilon_j=1.96\cdot\frac{\sqrt S}{N} \label{Eq.3} 
\end{align}
 
Scores outside the confidence interval will be considered out-of-bounds, and we will discard these scores. The mean opinion score(MOS) for the $j_{th}$ AIGI is calculated by the following formula:
\begin{align}
    MOS_j = \frac{1}{M} \sum_{i=1}^{M} r_{ij}^\prime \label{Eq.4} 
\end{align}

Here, $M$ represents the number of non-discarded scores, and $r_{ij}^\prime$ denote the rescaled non-discarded scores.
The final score for AIGIs is calculated by the following formula:
\begin{align}
     \text{Final}\_{\text{score}}  = MOS{\text{quality}} + MOS_{\text{authenticity}} + MOS_{\text{correspondence}}  \label{Eq.5} 
\end{align}

\subsection{Database Analysis}
To further demonstrate the evaluation of AI-generated images from the perspectives of quality, authenticity, and text-image correspondence, we present examples of high-quality AIGIs, low-quality AIGIs, high-authenticity AIGIs, low-authenticity AIGIs, high-text-image correspondence AIGIs, and low-text-image correspondence AIGIs as shown in Fig.3. Each evaluation perspective has its unique value. Fig.4 displays histograms of Mean Opinion Scores for quality, authenticity, text-image correspondence, and the final score, respectively. We can find that all the score distributions tend to be Gaussian distributions.

\begin{figure}[t]
\centering
\includegraphics[width=11cm,height=3.19cm]{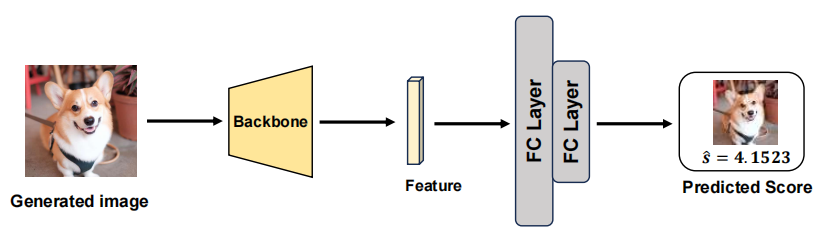} \label{1}
\caption{The pipeline of the proposed NR-AIGCIQA method. It contains two segments: image features extraction network and score regression network. The input images are fed into the image features extraction network to extract features. Then, we pass the feature to the score regression network composed of two fully connected layers to obtain the predicted score.}
\end{figure}
  
\section{Approach}
In this section, we present two AIGCIQA benchmark models for PKU-I2IQA database, encompassing NR-IQA method and FR-IQA method. Fig.5 and Fig.6 illustrate the pipelines for NR-AIGCIQA and FR-AIGCIQA methods, respectively.

\subsection{Problem Formulation}

For a given AIGI $I_g$ with score label $s$ , our proposed NR-AIGCIQA method first utilizes a visual backbone to extract features from the generated image. Subsequently, a regression network composed of two fully connected layers is employed to regress the predicted score. This method can be represented as:
\begin{align}
   \hat{s} = R_\theta(F_w(I_g)) \label{Eq.4} 
\end{align}

Here, $R_\theta$ and $F_w$ denote the regression network with parameters $\theta$ and the feature extraction network with parameters $w$,respectively.

For a given AIGI $I_g$ with score label $s$ and an image prompt $I_p$ , our proposed FR-AIGCIQA method first employs a shared-weights backbone network to extract features from $I_g$ and $I_p$, separately. These features are then fused using concatenation, and finally, a regression network composed of two fully connected layers is applied to regress the predicted score. This method can be represented as:
\begin{align}
   \hat{s} = R_\theta(\text{Concat}(F_w(I_g), F_w(I_p))) \label{Eq.4} 
\end{align}

Here, $R_\theta$ and $F_w$ denote the regression network with parameters $\theta$ and the feature extraction network with parameters $w$, respectively.

\subsection{Benchmark Model}
Due to the images in the PKU-I2IQA database being generated by image prompts and text prompts and each generated image corresponds to a specific image prompt, FR-IQA methods can be employed in this scenario. Additionally, we tested the NR-IQA methods on the PKU-I2IQA database which does not utilize prompt images as reference images during training and testing.

\begin{figure}[t]
\centering
\includegraphics[width=11cm,height=5.4cm]{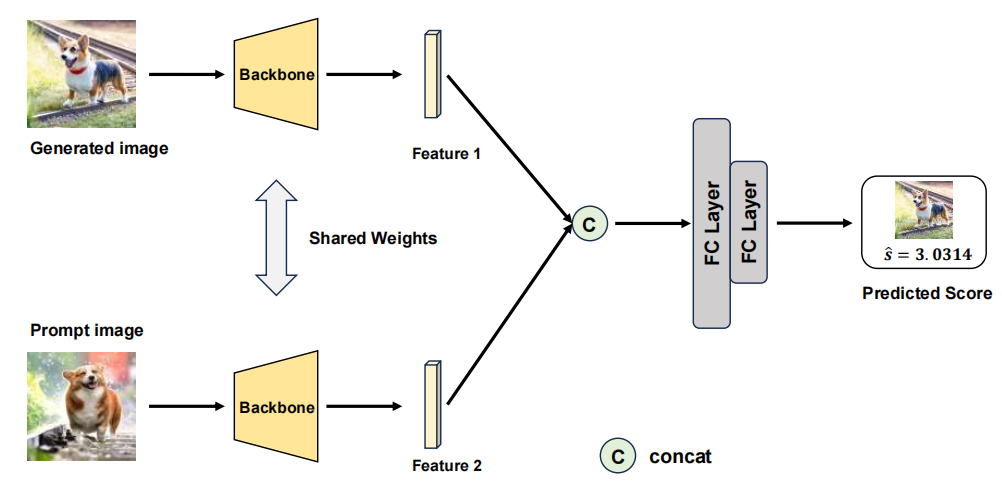} \label{1}
\caption{The pipeline of the proposed FR-AIGCIQA method. The generated image and prompt image are fed into a shared-weights vision backbone to extract features. Then the features are fused by concatenation. Finally, we pass the fused feature to the score regression network composed of two fully connected layers to obtain the predicted score.}
\end{figure}

Our proposed benchmark models based on the NR-IQA method and FR-IQA method consist of two components: a feature extraction network and a score regression network. We will provide detailed descriptions of these two components below.
\subsubsection{Feature Extraction Network.}
Initially, classical image quality assessment models relies on handcrafted feature-based methods. However, the advent of convolutional neural networks has led to the predominance of deep learning-based feature extraction, which surpasses traditional methods in performance. Deep learning approaches, unlike their handcrafted counterparts that rely on empirical rules, are data-driven and excel in extracting abstract and high-level semantic features from images. In our proposed NR-AIGCIQA method and FR-AIGCIQA method, we employ several backbone network models (VGG16\cite{simonyan2014vgg}, VGG19\cite{simonyan2014vgg}, ResNet18\cite{he2016resnet}, ResNet50\cite{he2016resnet}, and InceptionV4\cite{szegedy2017inception}) pre-trained on the ImageNet\cite{russakovsky2015imagenet} for feature extraction from input images.
\subsubsection{Score Regression Network.}
For the image features extracted by the backbone network with a feature dimension of (B, D), we employ a score regression network composed of two fully connected layers with dimensions $D \times \frac{D}{2}$  and  $\frac{D}{2} \times 1$ to regress the predictd score $\hat{s}$.
\subsubsection{Loss Function.}
We optimize the parameters of the feature extraction network and the score regression network by minimizing the mean squared error between the predicted score $\hat{s}$ and the true score $s$:
\begin{align}
   L_{MSE}(\theta, w | I) = ||\hat{s} - s||^2 \label{Eq.4} 
\end{align}
 
Here, the parameters $\theta$ and $w$ correspond to the parameters of the regression network and the feature extraction network, respectively.

\section{Experiment}
\subsection{Implementation Details}
Our experiments were conducted on the NVIDIA A40, using PyTorch 1.11.0 and CUDA 11.3 for both training and testing.

In the PKU-I2IQA database, scores are annotated across four dimensions: quality, authenticity, text-image correspondence, and a final score. To accurately evaluate model performance, we train individual models for each scoring category. For feature extraction from input images, we select several backbone network models pre-trained on the ImageNet\cite{russakovsky2015imagenet}, including VGG16\cite{simonyan2014vgg}, VGG19\cite{simonyan2014vgg}, ResNet18\cite{he2016resnet}, ResNet50\cite{he2016resnet}, and InceptionV4\cite{szegedy2017inception}.
Due to the inconsistency in input dimensions of the backbone networks such as InceptionV4 with the image sizes in our dataset, specific preprocessing is required. For InceptionV4, we adjust image sizes to 320$\times$320, followed by random cropping to 299$\times$299 and a 50\% chance of horizontal flipping. For the other networks, images are resized to 256$\times$256, then randomly cropped to 224$\times$224 with the same probability of horizontal flipping.
During training, the batch size $B$ is set to $8$. We utilize the Adam optimizer\cite{kingma2014adam} with a learning rate of $1 \times 10^{-4}$ and weight decay of $1 \times 10^{-5}$. The training loss employed is mean squared error (MSE) loss. In the testing phase, the batch size $B$ is set to $20$.

To evaluate the AIGI generative models in the PKU-I2IQA database, we split the data into training and test sets at a 3:1 ratio for each category produced by each generative model. We then report the performance of our two proposed methods alongside various pre-trained backbone networks.

We compare the performance of the following methods on the PKU-I2IQA database:

$\bullet$  \bm{$F^\ast+R$} (Baseline): Corresponds to the NR-AIGCIQA method. $\ast$ indicates that our model is trained and tested exclusively with AIGIs, without the use of any reference images.

$\bullet$  \bm{$F+R$}: Corresponds to the FR-AIGCIQA method. This method employ a combination of prompt images and AIGIS as inputs during both the training and testing phases of the model.

\subsection{Evaluation Criteria}
Following prior research \cite{zhang2023perceptual,wang2023aigciqa2023,lao2022attentions,yang2022maniqa}, we utilize the Spearman rank correlation coefficient (SRCC) and Pearson linear correlation coefficient (PLCC) as evaluation metrics to evaluate the performance of our model.

The SRCC is defined as follows:
\begin{align}
   \text{SRCC} = 1 - \frac{6 \sum_{i=1}^{N} d_i^2}{N(N^2 - 1)} \label{Eq.4} 
\end{align}

Here, $N$ represents the number of test images, and $d_i$ denotes the difference in ranking between the true quality scores and the predicted quality scores for the $i_{th}$ test image.

The PLCC is defined as follows:
\begin{align}
    \text{PLCC} = \frac{\sum_{i=1}^{N}(si - \mu_{s_i})(\hat{s}_i - \hat{\mu}_{s_i})}{\sqrt{\sum_{i=1}^{N}(s_i - \mu_{s_i})^2 \sum_{i=1}^{N}(\hat{s}_i - \hat{\mu}_{s_i})^2}}\label{Eq.4} 
\end{align}

Here, $s_i$ and $\hat{s}_i$ represent the true and predicted quality scores, respectively, for the $i_{th}$ image. $\mu_{s_i}$ and $\hat{\mu}_{s_i}$ are their respective means, and $N$ is the number of test images.
Both SRCC and PLCC are metrics used to evaluate the relationship between two sets of variables. They range between $-1$ and $1$, where a positive value indicates a positive correlation and a negative value indicates a negative correlation, and a larger value means a better performance.
\subsection{Results}
 The performance results of the proposed methods on the PKU-I2IQA database are exhibited in Table 2.

 \begin{table}[t]
\caption{Performance comparisons of the two proposed methods mentioned above on the proposed PKU-I2IQA database. $\ast$ indicates that our model is trained and tested exclusively with AIGIs, without the use of any reference images. The best performance results are marked in \textcolor{red}{RED} and the second-best performance results are marked in \textcolor{blue}{BLUE} .}
\centering
%\begin{tabularx}{\textwidth}
\begin{tabular}{@{}l|cccccccc@{}}
\toprule
 \multirow{2}{*}{\textbf{Method}} & \multicolumn{2}{c}{\textbf{Quality}} & \multicolumn{2}{c}{\textbf{Authenticity}} & \multicolumn{2}{c}{\textbf{Correspondence}} & \multicolumn{2}{c}{\textbf{Final\_score}} \\ \cline{2-9}
                       & \textbf{SRCC}       & \textbf{PLCC}      & \textbf{SRCC}        & \textbf{PLCC}       & \textbf{SRCC}          & \textbf{PLCC}       & \textbf{SRCC}        & \textbf{PLCC}       \\ \hline
\textbf{VGG16*}\cite{simonyan2014vgg}                  & 0.6734     & 0.6854    & 0.6449      & 0.6975     & 0.7130        & 0.7095     & 0.6831      & 0.7065     \\
\textbf{VGG19*}\cite{simonyan2014vgg}                 & 0.6836     & 0.6855    & 0.6352      & 0.6845     & 0.7383        & 0.7348     & 0.6889      & 0.7099     \\
\textbf{ResNet18*}\cite{he2016resnet}               & 0.6885     & 0.7112    & 0.6684      & 0.7108     & 0.7492        & 0.7317     & 0.7093      & 0.7252     \\
\textbf{ResNet50*}\cite{he2016resnet}               & 0.6876     & 0.6875    & 0.6530      & 0.6918     & 0.7456        & 0.7385     & 0.7272      & 0.7426     \\
\textbf{InceptionV4*}\cite{szegedy2017inception}            & 0.6988     & 0.7076    & \textcolor{blue}{0.6733}      & 0.7191     & 0.7509        & 0.7306     & 0.7221      & 0.7314     \\ \hline
\textbf{VGG16}\cite{simonyan2014vgg}                   & 0.6825     & 0.6918    & 0.6468      & 0.7005     & 0.7589        & 0.7740     & 0.7214      & 0.7478     \\
\textbf{VGG19}\cite{simonyan2014vgg}                   & 0.6832     & 0.7093    & 0.6505      & 0.7056     & 0.7594        & 0.7741     & 0.7084      & 0.7488     \\
\textbf{ResNet18}\cite{he2016resnet}                & \textcolor{red}{0.7063}     & \textcolor{red}{0.7249}    & 0.6724      & \textcolor{blue}{0.7220}     & \textcolor{red}{0.7737}        & \textcolor{red}{0.7892}     & 0.7241      & \textcolor{blue}{0.7565}     \\
\textbf{ResNet50}\cite{he2016resnet}                & 0.6885     & 0.6968    & 0.6567      & 0.6983     & \textcolor{blue}{0.7662}        & \textcolor{blue}{0.7803}     & \textcolor{red}{0.7359}      & \textcolor{red}{0.7606}     \\
\textbf{InceptionV4}\cite{szegedy2017inception}             & \textcolor{blue}{0.7017}     & \textcolor{blue}{0.7246}    & \textcolor{red}{0.6788}      & \textcolor{red}{0.7298}     & 0.7626        & 0.7627     & \textcolor{blue}{0.7282}      & 0.7529     \\ \bottomrule
\end{tabular}
\end{table}

Based on the results reported in the Table 2, we can draw several conclusions:

$\bullet$  The benchmark model of the FR-AIGCIQA method outperforms the benchmark model of NR-AIGCIQA method.

$\bullet$  Among the backbone networks we utilize, ResNet18\cite{he2016resnet} performs the best in terms of quality and correspondence on the PKU-I2IQA database. ResNet50\cite{he2016resnet} exhibits the best on Final\_score, while InceptionV4\cite{szegedy2017inception} demonstrates the best performance on authenticity.

$\bullet$  Overall, ResNet18\cite{he2016resnet} exhibits the best performance, followed by Inceptionv4\cite{szegedy2017inception} and ResNet50\cite{he2016resnet}.

\section{Conclusion}
In this paper, we first introduce an image-to-image database named PKU-I2IQA for AIGCIQA based on human perception. We select 200 categories from the well-known large-scale image database ImageNet in the field of computer vision and collecte corresponding images for each selected category as image prompts for generating images using different generative models. For each image prompt, we generate four images randomly for each model. Therefore, the PKU-I2IQA database comprises a total of 1600 images corresponding to 200 image prompts. We conduct a well-organized subjective experiment to collect quality labels for AIGIs and then conduct a comprehensive analysis of the PKU-I2IQA database.

 Furthermore, we propose two benchmark models, namely NR-AIGCIQA and FR-AIGCIQA. Finally, we conduct benchmark experiments and compare the performance of the proposed benchmark models alongside various pre-trained backbone networks. The results indicate the following: first, despite the proposed benchmark models exhibiting certain performance, there is still considerable room for improvement in designing AIGCIQA models; second, the benchmark model of the FR-AIGCIQA method outperforms the benchmark model of the NR-AIGCIQA method. Therefore, in future research, we will focus on how to introduce reference images in scenarios like text-to-image generation without image prompts to enhance the model's performance. Additionally, we conduct cross-model evaluation experiments. Specifically, we train our models on images generated by one AIGI model and test it on images generated by another. The results indicate that the proposed benchmark model exhibits weak generalization when evaluate different AIGI models. We do not include this part in the paper, and in the future, we aim to further research and design AIGCIQA models with stronger generalization capabilities.

\clearpage
\bibliographystyle{plain}   
\bibliography{I2IQA} 

\begin{thebibliography}{10}

\bibitem{Pixabay}
Pixabay.
\newblock \url{https://pixabay.com/}, 2010.

\bibitem{r1}
Midjourney.
\newblock \url{https://www.midjourney.com/home/}, 2022.

\bibitem{bao2023unidiff}
Fan Bao, Shen Nie, Kaiwen Xue, Chongxuan Li, Shi Pu, Yaole Wang, Gang Yue, Yue
  Cao, Hang Su, and Jun Zhu.
\newblock One transformer fits all distributions in multi-modal diffusion at
  scale.
\newblock {\em arXiv preprint arXiv:2303.06555}, 2023.

\bibitem{8063957}
Sebastian Bosse, Dominique Maniry, Klaus-Robert Müller, Thomas Wiegand, and
  Wojciech Samek.
\newblock Deep neural networks for no-reference and full-reference image
  quality assessment.
\newblock {\em IEEE Transactions on Image Processing}, 27(1):206--219, 2018.

\bibitem{chen2021full}
Chenglizhao Chen, Hongmeng Zhao, Huan Yang, Teng Yu, Chong Peng, and Hong Qin.
\newblock Full-reference screen content image quality assessment by fusing
  multilevel structure similarity.
\newblock {\em ACM Transactions on Multimedia Computing, Communications, and
  Applications (TOMM)}, 17(3):1--21, 2021.

\bibitem{dosovitskiy2020vit}
Alexey Dosovitskiy, Lucas Beyer, Alexander Kolesnikov, Dirk Weissenborn,
  Xiaohua Zhai, Thomas Unterthiner, Mostafa Dehghani, Matthias Minderer, Georg
  Heigold, Sylvain Gelly, et~al.
\newblock An image is worth 16x16 words: Transformers for image recognition at
  scale.
\newblock {\em arXiv preprint arXiv:2010.11929}, 2020.

\bibitem{c5}
Huiyu Duan, Xiongkuo Min, Yucheng Zhu, Guangtao Zhai, Xiaokang Yang, and
  Patrick Le~Callet.
\newblock Confusing image quality assessment: Toward better augmented reality
  experience.
\newblock {\em IEEE Transactions on Image Processing}, 31:7206--7221, 2022.

\bibitem{r3}
Huiyu Duan, Wei Shen, Xiongkuo Min, Yuan Tian, Jae-Hyun Jung, Xiaokang Yang,
  and Guangtao Zhai.
\newblock Develop then rival: A human vision-inspired framework for
  superimposed image decomposition.
\newblock {\em IEEE Transactions on Multimedia}, 25:4267--4281, 2023.

\bibitem{c3}
Huiyu Duan, Guangtao Zhai, Xiongkuo Min, Yucheng Zhu, Yi~Fang, and Xiaokang
  Yang.
\newblock Perceptual quality assessment of omnidirectional images.
\newblock In {\em 2018 IEEE International Symposium on Circuits and Systems
  (ISCAS)}, pages 1--5, 2018.

\bibitem{c4}
Huiyu Duan, Guangtao Zhai, Xiaokang Yang, Duo Li, and Wenhan Zhu.
\newblock Ivqad 2017: An immersive video quality assessment database.
\newblock In {\em 2017 International Conference on Systems, Signals and Image
  Processing (IWSSIP)}, pages 1--5, 2017.

\bibitem{r19}
Xinbo Gao, Fei Gao, Dacheng Tao, and Xuelong Li.
\newblock Universal blind image quality assessment metrics via natural scene
  statistics and multiple kernel learning.
\newblock {\em IEEE Transactions on Neural Networks and Learning Systems},
  24(12):2013--2026, 2013.

\bibitem{r20}
Deepti Ghadiyaram and Alan~C Bovik.
\newblock Perceptual quality prediction on authentically distorted images using
  a bag of features approach.
\newblock {\em Journal of vision}, 17(1):32--32, 2017.

\bibitem{goodfellow2014generative}
Ian Goodfellow, Jean Pouget-Abadie, Mehdi Mirza, Bing Xu, David Warde-Farley,
  Sherjil Ozair, Aaron Courville, and Yoshua Bengio.
\newblock Generative adversarial nets.
\newblock {\em Advances in neural information processing systems}, 27, 2014.

\bibitem{r6}
Ishaan Gulrajani, Faruk Ahmed, Martin Arjovsky, Vincent Dumoulin, and Aaron~C
  Courville.
\newblock Improved training of wasserstein gans.
\newblock {\em Advances in neural information processing systems}, 30, 2017.

\bibitem{he2016resnet}
Kaiming He, Xiangyu Zhang, Shaoqing Ren, and Jian Sun.
\newblock Deep residual learning for image recognition.
\newblock In {\em Proceedings of the IEEE conference on computer vision and
  pattern recognition}, pages 770--778, 2016.

\bibitem{hessel2021clipscore}
Jack Hessel, Ari Holtzman, Maxwell Forbes, Ronan~Le Bras, and Yejin Choi.
\newblock Clipscore: A reference-free evaluation metric for image captioning.
\newblock {\em arXiv preprint arXiv:2104.08718}, 2021.

\bibitem{r7}
Martin Heusel, Hubert Ramsauer, Thomas Unterthiner, Bernhard Nessler, and Sepp
  Hochreiter.
\newblock Gans trained by a two time-scale update rule converge to a local nash
  equilibrium.
\newblock {\em Advances in neural information processing systems}, 30, 2017.

\bibitem{ho2020ddpm}
Jonathan Ho, Ajay Jain, and Pieter Abbeel.
\newblock Denoising diffusion probabilistic models.
\newblock {\em Advances in neural information processing systems},
  33:6840--6851, 2020.

\bibitem{cnriqa}
Le~Kang, Peng Ye, Yi~Li, and David Doermann.
\newblock Convolutional neural networks for no-reference image quality
  assessment.
\newblock In {\em 2014 IEEE Conference on Computer Vision and Pattern
  Recognition}, pages 1733--1740, 2014.

\bibitem{kingma2014adam}
Diederik~P Kingma and Jimmy Ba.
\newblock Adam: A method for stochastic optimization.
\newblock {\em arXiv preprint arXiv:1412.6980}, 2014.

\bibitem{kingma2013vae}
Diederik~P Kingma and Max Welling.
\newblock Auto-encoding variational bayes.
\newblock {\em arXiv preprint arXiv:1312.6114}, 2013.

\bibitem{r4}
Yuval Kirstain, Adam Polyak, Uriel Singer, Shahbuland Matiana, Joe Penna, and
  Omer Levy.
\newblock Pick-a-pic: An open dataset of user preferences for text-to-image
  generation.
\newblock {\em arXiv preprint arXiv:2305.01569}, 2023.

\bibitem{lao2022attentions}
Shanshan Lao, Yuan Gong, Shuwei Shi, Sidi Yang, Tianhe Wu, Jiahao Wang, Weihao
  Xia, and Yujiu Yang.
\newblock Attentions help cnns see better: Attention-based hybrid image quality
  assessment network.
\newblock In {\em Proceedings of the IEEE/CVF conference on computer vision and
  pattern recognition}, pages 1140--1149, 2022.

\bibitem{r5}
Kimin Lee, Hao Liu, Moonkyung Ryu, Olivia Watkins, Yuqing Du, Craig Boutilier,
  Pieter Abbeel, Mohammad Ghavamzadeh, and Shixiang~Shane Gu.
\newblock Aligning text-to-image models using human feedback.
\newblock {\em arXiv preprint arXiv:2302.12192}, 2023.

\bibitem{AGIQA-3K}
Chunyi Li, Zicheng Zhang, Haoning Wu, Wei Sun, Xiongkuo Min, Xiaohong Liu,
  Guangtao Zhai, and Weisi Lin.
\newblock Agiqa-3k: An open database for ai-generated image quality assessment.
\newblock {\em IEEE Transactions on Circuits and Systems for Video Technology},
  pages 1--1, 2023.

\bibitem{liu2017rankiqa}
Xialei Liu, Joost Van De~Weijer, and Andrew~D Bagdanov.
\newblock Rankiqa: Learning from rankings for no-reference image quality
  assessment.
\newblock In {\em Proceedings of the IEEE international conference on computer
  vision}, pages 1040--1049, 2017.

\bibitem{liu2021swin}
Ze~Liu, Yutong Lin, Yue Cao, Han Hu, Yixuan Wei, Zheng Zhang, Stephen Lin, and
  Baining Guo.
\newblock Swin transformer: Hierarchical vision transformer using shifted
  windows.
\newblock In {\em Proceedings of the IEEE/CVF international conference on
  computer vision}, pages 10012--10022, 2021.

\bibitem{c6}
Xiongkuo Min, Kede Ma, Ke~Gu, Guangtao Zhai, Zhou Wang, and Weisi Lin.
\newblock Unified blind quality assessment of compressed natural, graphic, and
  screen content images.
\newblock {\em IEEE Transactions on Image Processing}, 26(11):5462--5474, 2017.

\bibitem{nichol2021glide}
Alex Nichol, Prafulla Dhariwal, Aditya Ramesh, Pranav Shyam, Pamela Mishkin,
  Bob McGrew, Ilya Sutskever, and Mark Chen.
\newblock Glide: Towards photorealistic image generation and editing with
  text-guided diffusion models.
\newblock {\em arXiv preprint arXiv:2112.10741}, 2021.

\bibitem{r8}
Mayu Otani, Riku Togashi, Yu~Sawai, Ryosuke Ishigami, Yuta Nakashima, Esa
  Rahtu, Janne Heikkil{\"a}, and Shin’ichi Satoh.
\newblock Toward verifiable and reproducible human evaluation for text-to-image
  generation.
\newblock In {\em Proceedings of the IEEE/CVF Conference on Computer Vision and
  Pattern Recognition}, pages 14277--14286, 2023.

\bibitem{radford2021clip}
Alec Radford, Jong~Wook Kim, Chris Hallacy, Aditya Ramesh, Gabriel Goh,
  Sandhini Agarwal, Girish Sastry, Amanda Askell, Pamela Mishkin, Jack Clark,
  et~al.
\newblock Learning transferable visual models from natural language
  supervision.
\newblock In {\em International conference on machine learning}, pages
  8748--8763. PMLR, 2021.

\bibitem{dalle}
Aditya Ramesh, Prafulla Dhariwal, Alex Nichol, Casey Chu, and Mark Chen.
\newblock Hierarchical text-conditional image generation with clip latents.
\newblock {\em arXiv preprint arXiv:2204.06125}, 1(2):3, 2022.

\bibitem{r2}
Robin Rombach, Andreas Blattmann, Dominik Lorenz, Patrick Esser, and Bj{\"o}rn
  Ommer.
\newblock High-resolution image synthesis with latent diffusion models.
\newblock In {\em Proceedings of the IEEE/CVF conference on computer vision and
  pattern recognition}, pages 10684--10695, 2022.

\bibitem{russakovsky2015imagenet}
Olga Russakovsky, Jia Deng, Hao Su, Jonathan Krause, Sanjeev Satheesh, Sean Ma,
  Zhiheng Huang, Andrej Karpathy, Aditya Khosla, Michael Bernstein, et~al.
\newblock Imagenet large scale visual recognition challenge.
\newblock {\em International journal of computer vision}, 115:211--252, 2015.

\bibitem{fr3}
Soomin Seo, Sehwan Ki, and Munchurl Kim.
\newblock A novel just-noticeable-difference-based saliency-channel attention
  residual network for full-reference image quality predictions.
\newblock {\em IEEE Transactions on Circuits and Systems for Video Technology},
  31(7):2602--2616, 2021.

\bibitem{simonyan2014vgg}
Karen Simonyan and Andrew Zisserman.
\newblock Very deep convolutional networks for large-scale image recognition.
\newblock {\em arXiv preprint arXiv:1409.1556}, 2014.

\bibitem{bmsb}
Wei Sun, Huiyu Duan, Xiongkuo Min, Li~Chen, and Guangtao Zhai.
\newblock Blind quality assessment for in-the-wild images via hierarchical
  feature fusion strategy.
\newblock In {\em 2022 IEEE International Symposium on Broadband Multimedia
  Systems and Broadcasting (BMSB)}, pages 01--06, 2022.

\bibitem{szegedy2017inception}
Christian Szegedy, Sergey Ioffe, Vincent Vanhoucke, and Alexander Alemi.
\newblock Inception-v4, inception-resnet and the impact of residual connections
  on learning.
\newblock In {\em Proceedings of the AAAI conference on artificial
  intelligence}, volume~31, 2017.

\bibitem{szegedy2015google}
Christian Szegedy, Wei Liu, Yangqing Jia, Pierre Sermanet, Scott Reed, Dragomir
  Anguelov, Dumitru Erhan, Vincent Vanhoucke, and Andrew Rabinovich.
\newblock Going deeper with convolutions.
\newblock In {\em Proceedings of the IEEE conference on computer vision and
  pattern recognition}, pages 1--9, 2015.

\bibitem{ITUR}
I.~T. Union.
\newblock Methodology for the subjective assess- ment of the quality of
  television pictures.
\newblock {\em ITU-R Recommendation BT. 500-11}, 2002.

\bibitem{wang2023aigciqa2023}
Jiarui Wang, Huiyu Duan, Jing Liu, Shi Chen, Xiongkuo Min, and Guangtao Zhai.
\newblock Aigciqa2023: A large-scale image quality assessment database for ai
  generated images: from the perspectives of quality, authenticity and
  correspondence.
\newblock {\em arXiv preprint arXiv:2307.00211}, 2023.

\bibitem{mdaqa}
Tao Wang, Wei Sun, Xiongkuo Min, Wei Lu, Zicheng Zhang, and Guangtao Zhai.
\newblock A multi-dimensional aesthetic quality assessment model for mobile
  game images.
\newblock In {\em 2021 International Conference on Visual Communications and
  Image Processing (VCIP)}, pages 1--5, 2021.

\bibitem{yang2022maniqa}
Sidi Yang, Tianhe Wu, Shuwei Shi, Shanshan Lao, Yuan Gong, Mingdeng Cao, Jiahao
  Wang, and Yujiu Yang.
\newblock Maniqa: Multi-dimension attention network for no-reference image
  quality assessment.
\newblock In {\em Proceedings of the IEEE/CVF Conference on Computer Vision and
  Pattern Recognition}, pages 1191--1200, 2022.

\bibitem{nr-iqa}
Peng Ye and David Doermann.
\newblock No-reference image quality assessment using visual codebooks.
\newblock {\em IEEE Transactions on Image Processing}, 21(7):3129--3138, 2012.

\bibitem{zhang2023addingcontrolnet}
Lvmin Zhang, Anyi Rao, and Maneesh Agrawala.
\newblock Adding conditional control to text-to-image diffusion models.
\newblock In {\em Proceedings of the IEEE/CVF International Conference on
  Computer Vision}, pages 3836--3847, 2023.

\bibitem{zhang2023perceptual}
Zicheng Zhang, Chunyi Li, Wei Sun, Xiaohong Liu, Xiongkuo Min, and Guangtao
  Zhai.
\newblock A perceptual quality assessment exploration for aigc images.
\newblock {\em arXiv preprint arXiv:2303.12618}, 2023.

\bibitem{zhou2022lafite}
Yufan Zhou, Ruiyi Zhang, Changyou Chen, Chunyuan Li, Chris Tensmeyer, Tong Yu,
  Jiuxiang Gu, Jinhui Xu, and Tong Sun.
\newblock Towards language-free training for text-to-image generation.
\newblock In {\em Proceedings of the IEEE/CVF Conference on Computer Vision and
  Pattern Recognition}, pages 17907--17917, 2022.

\bibitem{MetaIQA}
Hancheng Zhu, Leida Li, Jinjian Wu, Weisheng Dong, and Guangming Shi.
\newblock Metaiqa: Deep meta-learning for no-reference image quality
  assessment.
\newblock In {\em 2020 IEEE/CVF Conference on Computer Vision and Pattern
  Recognition (CVPR)}, pages 14131--14140, 2020.

\end{thebibliography}
\end{document}